\documentclass[lettersize, journal]{IEEEtran}
\usepackage{amsmath,amsfonts}
\usepackage{algorithmic}
\usepackage{algorithm}
\usepackage{array}
\usepackage{booktabs}
\usepackage[caption=false,font=footnotesize,labelfont=rm,textfont=rm]{subfig}
\usepackage{textcomp}
\usepackage{stfloats}
\usepackage{url}
\usepackage{verbatim}
\usepackage{graphicx}
\usepackage{subfloat}
\usepackage{hyperref}
\usepackage{bm}
\usepackage{multirow}
\usepackage{cite}
\usepackage{dsfont}
\usepackage{soul}
\usepackage{xcolor}

\usepackage{lineno}

\soulregister{\cite}{7}  % 例如，注册\cite 命令
\soulregister{\ref}{7}
\soulregister{\hyperref}{7}

\hypersetup{
	colorlinks=true,
	linkcolor=blue,
	filecolor=black,      
	urlcolor=black,
	citecolor=blue,
}
\usepackage{balance}

\usepackage{xcolor}

\begin{document}	
	
	\title{An Efficient Learning Control Framework with Sim-to-Real for String-Type Artificial Muscle-Driven Robotic Systems}
	
	\author{Jiyue Tao, Yunsong Zhang, Sunil Kumar Rajendran, and Feitian Zhang
		% <-this % stops a space
		\thanks{This work was partially supported by the National Key Research and Development Program of China under Grant 2022YFB4701900. \emph{(Corresponding author: Feitian Zhang.)}}
		\thanks{Jiyue Tao, Yunsong Zhang, and Feitian Zhang are with the Robotics and Control Laboratory, Department of Advanced Manufacturing and Robotics, College of Engineering, and the State Key Laboratory of Turbulence and Complex Systems, Peking University, Beijing, 100871, China (\href{mailto: jiyuetao@pku.edu.cn}{email: jiyuetao@pku.edu.cn}; \href{mailto: zhangyunsong@stu.pku.edu.cn}{email: zhangyunsong@stu.pku.edu.cn}; \href{mailto:feitian@pku.edu.cn}{email: feitian@pku.edu.cn}).}% <-this % stops a space
		\thanks{Sunil Kumar Rajendran is with the BSS Technologies Inc., Takoma Park, MD 20912, USA (\href{mailto: srajend2@gmu.edu}{email: srajend2@gmu.edu}).}
	}
	
	\markboth{}
	{Shell \MakeLowercase{\textit{et al.}}: A Sample Article Using IEEEtran.cls for IEEE Journals}
	
	
	\maketitle

	\begin{abstract}
		Robotic systems driven by artificial muscles present unique challenges due to the nonlinear dynamics of actuators and the complex designs of mechanical structures. Traditional model-based controllers often struggle to achieve desired control performance in such systems. Deep reinforcement learning (DRL), a trending machine learning technique widely adopted in robot control, offers a promising alternative. However, integrating DRL into these robotic systems faces significant challenges, including the requirement for large amounts of training data and the inevitable sim-to-real gap when deployed to real-world robots. This paper proposes an efficient reinforcement learning control framework with sim-to-real transfer to address these challenges. Bootstrap and augmentation enhancements are designed to improve the data efficiency of baseline DRL algorithms, while a sim-to-real transfer technique, namely randomization of muscle dynamics, is adopted to bridge the gap between simulation and real-world deployment. Extensive experiments and ablation studies are conducted utilizing two string-type artificial muscle-driven robotic systems including a two degree-of-freedom robotic eye and a parallel robotic wrist, the results of which demonstrate the effectiveness of the proposed learning control strategy.
	\end{abstract}
	
	
	\section{Introduction}
	\IEEEPARstart{A}{rtificial} muscles, designed to mimic biological counterparts, generate mechanical forces and movements in response to external stimuli like electrical or chemical signals. Well known for their high power-to-weight ratios and inherent compliance\cite{Muscle_survey}, artificial muscles offer a unique combination of strength and flexibility conducive to versatile and adaptive robot designs. Over the past decade, robots driven by artificial muscles have achieved significant advancements, demonstrating impressive performance across diverse domains, including but not limited to biomimetic robots\cite{Biomimetic_robot2, Biomimetic_robot3}, robot manipulators\cite{Robot_manipulator1, Robot_manipulator2}, and soft robots\cite{Soft_robot1, Soft_robot2}.
	\par
\begin{figure}[t]
		\centering
		\includegraphics[width=0.90\linewidth]{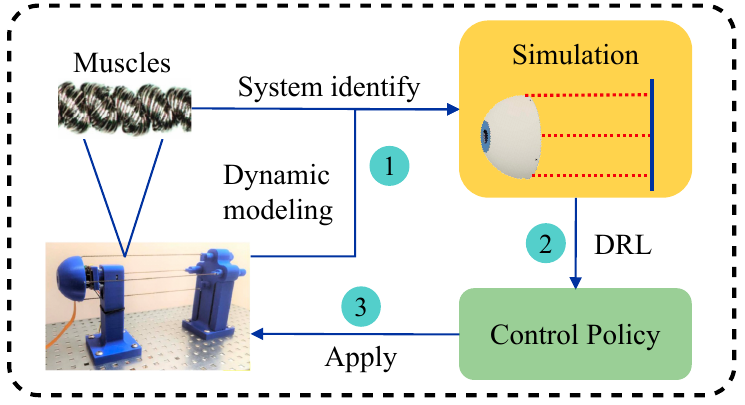}
		\caption{Schematic of standard procedure of applying DRL control in artificial muscle-driven robotic systems.}
		\label{fig:rl_solution}
	\end{figure}

	Previous research efforts have primarily centered around structural design and open-loop motion performance evaluation\cite{Biomimetic_robot2, Biomimetic_robot3, Robot_manipulator1, Robot_manipulator2, Soft_robot1, Soft_robot2}. Despite their potential, achieving accurate control of string-type artificial muscle-driven robots remains challenging due to their highly nonlinear and complex dynamics. Traditional control strategies, such as proportional-integral-derivative (PID) control\cite{PID1, PID2, PID3}, energy-based nonlinear control\cite{energybased}, and sliding mode control\cite{slidingmode1, slidingmode2} have been investigated to mitigate these challenges. Although these approaches have demonstrated success in specific tasks, a comprehensive, unified control strategy for string-type artificial muscle-driven robots is still lacking. As deep reinforcement learning (DRL) rapidly develops, learning-based controllers\cite{schlagenhauf2018control, you2017model, jiyue2024acc, Eye_RAL, rajendran2022design} have emerged as a promising alternative, offering the potential to overcome the limitations of conventional methods. The standard procedure of applying DRL in the targeted robotic systems typically follows the schematic illustrated in Fig.~\ref{fig:rl_solution}. Initially, identifying the relevant parameters and establishing a simulation environment enable a safe and efficient training process. The DRL algorithms are then trained within this simulation environment to develop an effective control policy. Finally, the learned policy is transferred to the physical robotic platform to achieve desired control performance. 
	\par	
	However, several challenges persist in the applications of DRL to string-type artificial muscle-driven robots. One major issue is the sim-to-real gap, which arises from inaccuracies in dynamic models, system identification errors, and variability in muscle string parameters\cite{zhang2017modeling, pawlowski2018modeling}. These factors often result in control strategies that demonstrate strong performance in simulation but struggle to transfer effectively to real-world applications. Another challenge is the low learning efficiency of DRL algorithms, which require substantial amounts of training data and iterations due to the complexity of neural networks and the trial and error learning process\cite{DRL_survey}. Addressing these two challenges is crucial for advancing control technologies in artificial muscle-driven robots.
	\par
	This paper presents a DRL-based control strategy designed to enhance the data efficiency and real-world applicability of controllers for artificial muscle-driven robots. The proposed framework is evaluated on two robotic systems actuated by artificial muscle strings: a biomimetic robotic eye driven by super-coiled polymers (SCPs)\cite{Eye_prot, SCP} and a parallel robotic wrist actuated by twisted-coiled actuators (TCAs)\cite{CTCA, TCA}. Specifically, we introduce a novel augmentation method for state-based data, which transforms a single transition data tuple into several, and we leverage an easy-to-implement PID controller to generate trajectories for early-stage training of DRL agents. Compared to the baseline algorithm (i.e., Soft Actor Critic~\cite{SAC}), the required training data is reduced by approximately $60\%$ in the targeted robotic systems.  Additionally, we leverage a zero-shot sim-to-real transfer technique on artificial muscle dynamics. The experimental results demonstrate a significant improvement in real-world control performance with the averaged error reduced by $36.3\%$.
	\par
	The main contributions of this paper are threefold. First, we enhance the training of DRL by leveraging PID controller-generated trajectories to accelerate early-stage training and introducing a novel state-based augmentation method that transforms single data transitions into multiple augmented samples. Second, we propose a zero-shot sim-to-real transfer technique specifically tailored for artificial muscle dynamics, effectively addressing the sim-to-real gap caused by dynamic model inaccuracies and parameter variability. Third, we evaluate our approach through extensive experiments on two distinct robotic systems, demonstrating significant reductions in both training data requirements and real-world control errors. These results validate the effectiveness of the proposed method and highlight its potential applicability to similar systems.
	
	%	\par
	%	The remainder of this paper is organized as follows. Section~\ref{related work} provides a review of the relevant literature. Section~\ref{preliminaries} introduces the basic concepts of applying reinforcement learning to the robotic systems studied. Section~\ref{experimental platforms} offers a concise overview of the robotic platforms utilized in this study. In Section~\ref{learning control framework}, we present a detailed description of our proposed efficient learning control framework. Section~\ref{experiment in simulation} and Section~\ref{real world experiments} discuss the experimental results, including training outcomes in simulation and control performances in real-world scenarios, respectively. Finally, concluding remarks and current limitations are discussed in Section~\ref{conclusions}.
	
	\section{Related Work}
	\label{related work}
	The control of artificial muscle-driven robotic systems using learning-based methods has garnered significant attention in recent years\cite{giorelli2015neural, yang2018soft, Eye_RAL, jiyue2024acc, kim2020learning, rajendran2022design, you2017model, thuruthel2018model}. Among these approaches, DRL has demonstrated effectiveness in handling diverse and complex tasks, positioning it as a promising control solution for a wide range of applications. An increasing number of researchers have leveraged DRL algorithms to address control challenges in artificial muscle-driven robots, spanning applications ranging from dexterous grasping of robotic hands\cite{schlagenhauf2018control} to agile motion control of soft robotic manipulators\cite{thuruthel2018model, you2017model}, and other related applications\cite{kim2020learning, rajendran2022design}. Many of these studies have noted that the low sample efficiency often leads to prolonged training time\cite{Eye_RAL, rajendran2022design}, but few have focused specifically on improving training efficiency. Several existing state-of-the-art (SOTA) data efficient reinforcement learning (RL) methods may be applicable\cite{REDQ, DroQ, RLPD, XQL}, but they have not been evaluated within the context of the robotic systems we are addressing. The unique nonlinear and slow response characteristics of artificial muscle may present challenges that are not adequately addressed by these existing methods. Our previous work\cite{jiyue2024acc} explored low learning efficiency problem but was limited to simulation-based experiments.
	\par
	Due to the sim-to-real gap, control strategies that perform well in simulation may not seamlessly transfer to real-world applications\cite{thuruthel2018model, you2017model}. Despite being labor-intensive, some researchers\cite{yang2018soft, kim2020learning} have attempted to directly train DRL algorithms on real robots. However, the artificial muscles considered in this paper are all thermally actuated, which inherently leads to energy inefficiency and limited cycling rates\cite{ mirvakili2014simple}. Continuous rapid transitions between heating and cooling often cause irreversible damage to the muscles, making direct training on real robots unsafe. Many researchers\cite{zhang2017modeling, pawlowski2018modeling, karami2020modeling} have proposed more accurate yet significantly more complex dynamic models to bridge the sim-to-real gap of artificial muscles. In this paper, however, we aim to directly improve the performance of DRL-based controllers by employing a sim-to-real transfer technique, thereby reducing the need for complicated mathematical modeling and precise parameter identification for artificial muscles.
	
	\section{Preliminaries}
	\label{preliminaries}
	In this section, we model the control problem of targeted robotic systems as a reinforcement learning problem by Markov Decision Process (MDP), which consists of the tuple $(\mathcal{S}, \mathcal{A}, \mathcal{R}, f, \gamma)$.
	
	\begin{itemize}
		\item $\mathcal{S}$ is the \textit{state space}. The state vector $s_t\in \mathcal{S}$ generally includes the current state variables of the robotic system and the target positions and/or orientations. In this paper, the motion state variables of the robotic systems are collectively defined as vector $\boldsymbol{x}$. The system output vector, containing the variables to be controlled, is denoted by $\boldsymbol{y}$. Correspondingly, the target system output is denoted by $\boldsymbol{y}^*$. In this way, the state vector $s_t$ is represented as $s_t=(\boldsymbol{x}_t, \boldsymbol{y}^*)$.
		
		\item $\mathcal{A}$ is the \textit{action space}, where each action $a_t \in \mathcal{A}$ is generated by the DRL agent based on the current state $s_t$ and its own policy. 
		
		\item $\mathcal{R}$ is the \textit{reward function} that maps the current state $s_t$ and action $a_t$ to an immediate reward $r_t=\mathcal{R}(s_t,a_t)$. In this paper, we develop a generalized quadratic-form reward function designed for the robotic systems of interest:
		\begin{equation}
			\label{reward function structure}
			r_t=-(e^\top_t Q_e e_t +a^\top_tR_a a_t)+ r_{\text{bonus}},
		\end{equation}
		where $e_t=|\boldsymbol{y}^* - \boldsymbol{y}|$ is the system tracking error, and $Q_e$ and $R_a$ represent the output and control weight matrices, respectively. These weights are determined based on empirical considerations with diagonal elements set to ensure terms in the reward function have comparable magnitudes. A bonus reward $r_{\text{bonus}}$ is given when the tracking error falls below a sufficiently small threshold. 
		
		\item $f$ represents the \textit{state transition function} and $\gamma$ is the \textit{discount factor}, which is set to $0.99$ in our experiments.
	\end{itemize}
	
	\begin{figure}[t]
		\centering
		\includegraphics[width=0.95\linewidth]{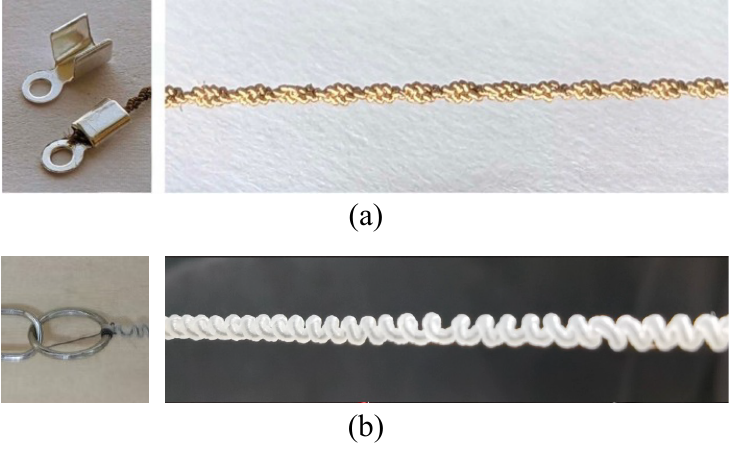}
		\vspace{-0.15in}
		\caption{Illustrations of SCP and TCA muscles. (a) Crimp fixtures for the SCP muscle ends (left), and a close-up shot of the fabricated SCP muscle (right); and (b) conductive ring with SMA wire wrapped around for the TCA muscle ends (left), and a close-up shot of the fabricated TCA muscle (right).}
		\label{fig:muscles}
	\end{figure}
	
	\section{Experimental Platforms}
	\label{experimental platforms}
\begin{figure}[t]
		\centering
		\includegraphics[width=0.9\linewidth]{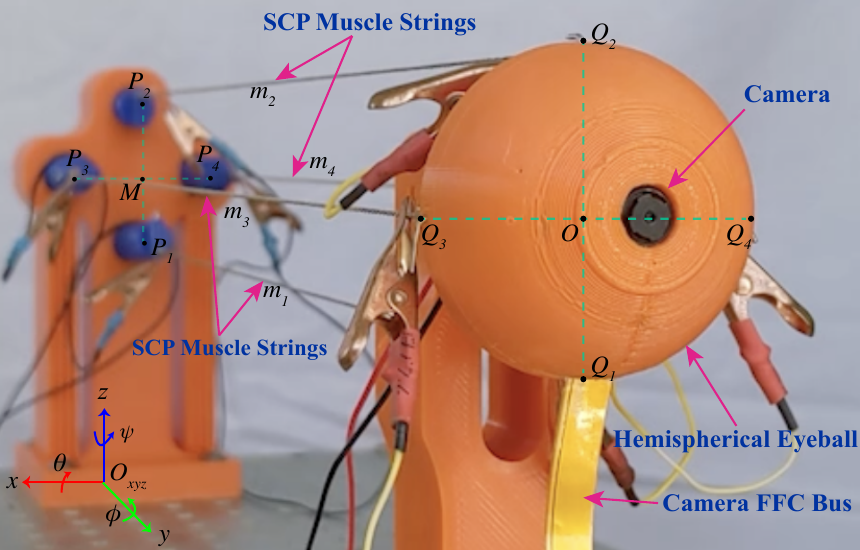}
		\caption{The 2-DOF robotic eye prototype driven by two pairs of SCP muscles.}
		\label{fig:eye}
	\end{figure}

	This section presents an overview of the actuators and robotic platforms used in this paper, including SCP and TCA artificial muscles (Fig.~\ref{fig:muscles}), a two-degrees-of-freedom (DOF) robotic eye (Fig.~\ref{fig:eye}) and a parallel robotic wrist (Fig.~\ref{fig:wrist}). For each system, we briefly describe the hardware design, state space, action space, and reward function design.
	
	\subsection{Artificial Muscles}
	Numerous fabrication methods for SCP and TCA muscles have been investigated in the literature\cite{Eye_prot, SCP, CTCA, TCA}. Specifically, for our robotic eye design, the SCP actuator is fabricated by sequentially twisting, coiling, and double-backing using a 145\,cm long silver-plated conductive nylon thread (Statex's Shieldex 117/17 dtex 2 ply HC+B). As illustrated in Fig.~\hyperref[fig:muscles]{\ref*{fig:muscles}(a)}, the ends of the SCP muscles are crimped using fold-over silver/copper bead crimps. They are connected to N-Channel enhancement MOSFETs (FQP30N06L) which provide electric power to the muscles. Application of voltage induces Joule heating on the actuator which causes the coils to expand laterally, thus generating contracting force.
	\par
	Regarding our TCA muscles, the Shape Memory Alloy (SMA) wires (0.18~mm diameter) serve as the conductive threads and skeletons. Highly elastic spandex fibers (creora spandex, power fit, 840~denier) are tightly wound around the SMAs, and the resulting composite fibers are then twisted into a coil configuration. As shown in Fig.~\hyperref[fig:muscles]{\ref*{fig:muscles}(b)}, both ends of the SMA wire are wrapped around conductive rings, which are also connected to MOSFETs. When voltage potential is applied to the rings, the skeleton generates Joule heat and the surrounding spandex fibers will contract in longitudinal direction, which leads to the untwisting of the compound twisted fiber structure. The TCA's coiled configuration converts this untwisting motion into linear contraction. Additionally, the SMA wire not only provides Joule heating, but also contributes additional contraction force owing to its shape memory effect.
	
	\subsection{Robotic Eye}
	The high occurrence of unsuccessful corrective surgeries for ocular motor disorders underscores the need for a muscle-driven robotic eye platform to assist ophthalmologists in gaining deeper insights into the biomechanics of human eye movement. In our previous work\cite{Eye_prot}, we proposed a prototype of a 2-DOF robotic eye and derived the geometric and dynamic models specific to this prototype. As illustrated in Fig.~\ref{fig:eye}, the hemispherical eyeball is driven by two pairs of SCP muscle actuators, accommodating a Raspberry Pi Mini Camera (RPiMC). Here, $Q_i$ represents the anchor position of muscle $m_i$ on the eyeball, through which the force generated by SCP muscle is transmitted. $P_i$ denotes the fixed endpoint of muscle $m_i$. The primary control objective is to precisely and rapidly adjust the eyeball's orientation towards the desired target angles, namely foveation control in eye movement study.
	
	\par
	The eyeball's center of rotation, denoted as $\boldsymbol{O}\in\mathbb{R}^3$, remains fixed at the origin within the reference frame $O_\mathrm{xyz}$. Here, the axes $\mathrm{x}$, $\mathrm{y}$, and $\mathrm{z}$ correspond to the transverse, sagittal, and vertical axes of the eyeball, respectively. Rotations around these axes are expressed as pitch $\theta$, roll $\phi$, and yaw $\psi$. The robotic eye system's state variables are represented as
	\begin{equation}
		\label{eq:eye-states}
		\boldsymbol{x}=[\theta,\dot{\theta},\psi,\dot{\psi},\Delta T_1,\Delta T_2,\Delta T_3,\Delta T_4]^\top,
	\end{equation}
	where  $\theta,\dot{\theta},\psi,\dot{\psi},$ represent the pitch, yaw angles, and their respective angular velocities for control. $\Delta T_i$ represents the temperature difference between the ambient temperature $T_\text{amb}$ and the current temperature $T_i$ of muscle $m_i$.
	
	\begin{figure}[t]
		\centering
		\includegraphics[width=0.9\linewidth]{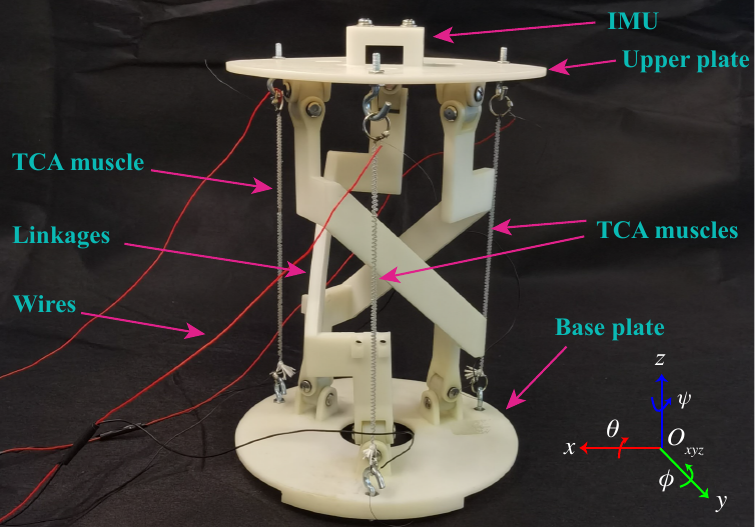}
		\caption{The parallel robotic wrist prototype actuated by three TCA muscles.}
		\label{fig:wrist}
	\end{figure}
	
	\par
	Focusing primarily on the robotic eye's orientation, the system's output vector is denoted as $\boldsymbol{y}=[\psi,\dot{\psi},\theta,\dot{\theta}]^\top$. Due to the difficulty and poor accuracy of measuring the temperature of each muscle in real time, the state vector of the DRL agent is defined by
	\begin{equation}
		s_t = [\theta_t,\dot{\theta}_t,\psi_t,\dot{\psi}_t,\theta^*,\psi^*]^\top.
	\end{equation}
	Here, $\theta^*$ and $\psi^*$ represent the target angles. The Euler angles and corresponding angular velocities are measured through visual feedback from the embedded RPiMC. In this paper, the actuating voltage applied to muscle $m_i$, denoted as $V_i$, is treated as a continuous variable within the defined range. Given the antagonistic configuration of the muscle pairs, the control input is defined by a constrained action vector, i.e.,
	\begin{equation}
		\begin{aligned}
			a_t=[&-V_1+V_2,-V_3+V_4]^\top  \\
			\text{s.t.}\quad& V_1\cdot V_2 = V_3 \cdot V_4 = 0,\ 0 \leq V_i \leq 10.
		\end{aligned}
	\end{equation}
	The constraint requires that only one muscle is activated at each time step in the pairs $(m_1,m_2)$ and $(m_3,m_4)$. The reward function follows the structure of \eqref{reward function structure}, where the output and control weight matrices are set to $Q_e = \text{diag}\{0.05, 0.25, 0.05, 0.25\}$ and $R_a = \text{diag}\{0.01, 0.01\}$, respectively. The bonus reward is given by
	\begin{equation}
		\label{eq:bonus1}
		r_{\text{bonus}}=2\times \left( \mathds{1}\{|\psi-\psi^*|<0.3\}+\mathds{1}\{|\theta-\theta^*|<0.3\} \right).
	\end{equation}
	Here, $\mathds{1}\{\cdot\}$ denotes the binary indicator function that yields a value of $1$ if the given condition holds true, and $0$ otherwise. 
	
	\subsection{Robotic Wrist}
	Research into artificial muscle-driven robotic wrist joints is of great importance to several domains, including the development of dexterous robotic systems with human-like capabilities, advancements in rehabilitation and prosthetics technology, and the enhancement of robotic manipulation versatility across various fields. We design a parallel robotic wrist prototype driven by three equally spaced compound TCAs\cite{CTCA} as illustrated in Fig.~\ref{fig:wrist}. The base plate remains stationary, while the upper plate, also known as the end effector, moves via three muscle strings. Linkages between the two plates are constructed to support a two degree-of-freedom motion of the upper plate\cite{Wrist_mechanism}. Given this configuration, we develop a simplified dynamics model under the assumption of a fixed rotational center for the upper plate. Similar to the foveation control of the robotic eye, the control objective entails precise regulation of the moving upper plate's orientation.
	\par
	The reference frame $O_\mathrm{xyz}$ is fixed on the moving upper plate, with the $\mathrm{z}$-axis perpendicular to the upper plate and pointing outward. $\theta,\phi,$ and $\psi$ represent the pitch, roll, and yaw angles, respectively. Similar to the robotic eye, the system state variables are defined as
	\begin{equation}
		\label{eq:wrist-states}
		\boldsymbol{x}=[\theta,\dot{\theta},\phi,\dot{\phi},\Delta T_1,\Delta T_2,\Delta T_3]^\top,
	\end{equation}
	where $\dot{\theta}$ and $\dot{\phi}$ represent the corresponding angular velocities, and $\Delta T_i$ is the temperature difference of muscle $m_i$ with respect to the ambient environment, with $i\in\{1,2,3\}$. The system output vector is represented as $\boldsymbol{y}=[\theta,\dot{\theta},\phi,\dot{\phi}]^\top$. Likewise, the muscle's temperature is not used as an input for the DRL agent, and thus, the state vector is  defined as
	\begin{equation}
		s_t=[\theta_t,\dot{\theta}_t,\phi_t,\dot{\phi}_t,\theta^*,\phi^*]^\top,
	\end{equation}
	where $\theta^*$ and $\phi^*$  denote the target angles, with the respective Euler angles and angular velocities measured using the inertial measurement unit (IMU) positioned at the center of the upper plate. The input action vector $a_t$ of the system is defined as:
	\begin{equation}
		a_t = [V_1, V_2, V_3]^\top,\ 0 \leq V_i \leq 10.
	\end{equation}
	The weight matrices are set to $Q_e=\text{diag}\{0.05, 0.2, 0.05, 0.2\}$ and $R_a=\text{diag}\{0.01, 0.01, 0.01\}$, respectively. The bonus reward is given by
	\begin{equation}
		\label{eq:bonus2}
		r_{\text{bonus}}=2\times \left( \mathds{1}\{|\theta-\theta^*|<0.5\}+\mathds{1}\{|\phi-\phi^*|<0.5\} \right).
	\end{equation}
	The threshold values in (5) and (9) are selected empirically based on testing experiments. Lower values may result in slower convergence and lower training stability due to reduced positive reinforcement, while higher values could diminish the precision of the learned control policy.
	
	\section{Learning Control Framework}
	\label{learning control framework}
	
	\begin{figure*}[t]
		\centering
		\includegraphics[width=0.90\linewidth]{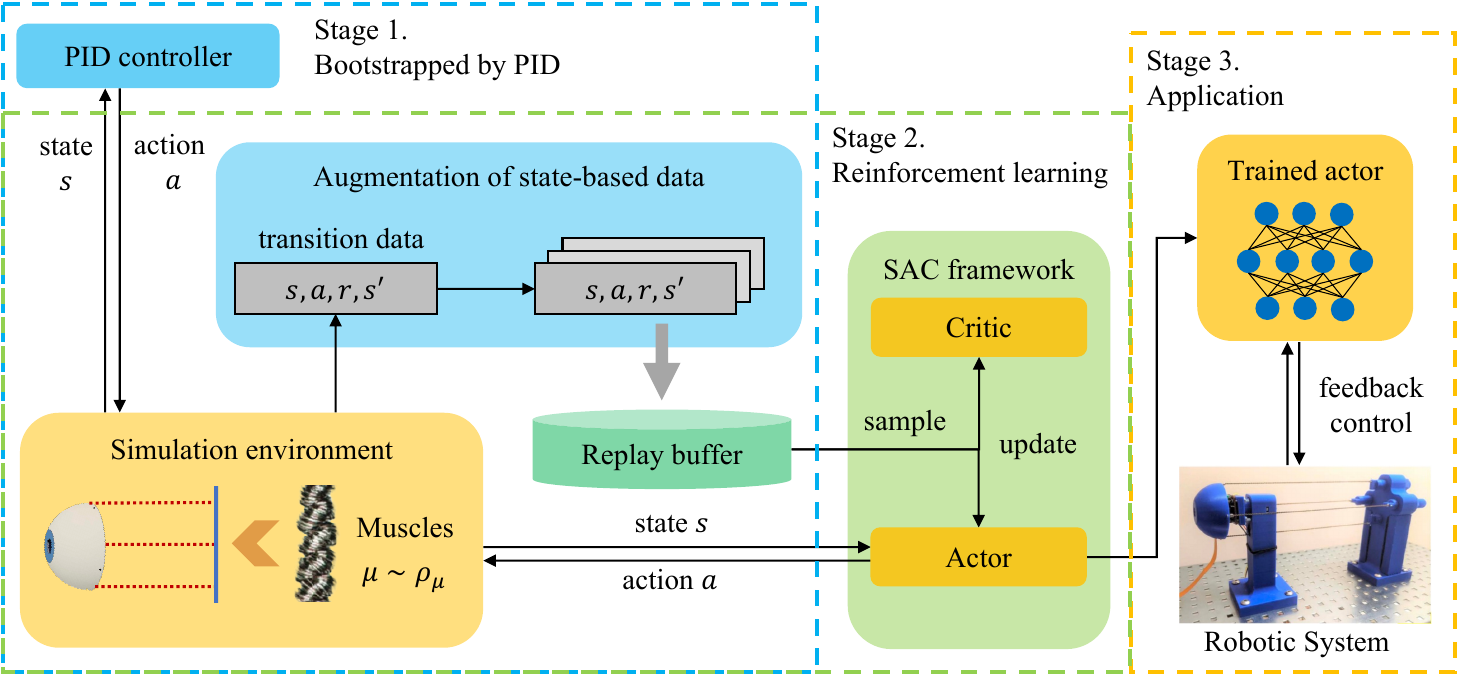}
		\caption{Overview of the proposed efficient learning control framework, including a two-stage training process in simulation and real-world application. Stage 1 utilizes bootstrapping with PID control, aiming at enhancing data efficiency. Stage 2 integrates the baseline SAC learning algorithm with state-based data augmentation and randomization of muscle dynamics for expediting training convergence and bridging sim-to-real gap, respectively.}
		\label{fig:framework}
	\end{figure*}
	This section provides a comprehensive overview of the proposed learning control framework, as illustrated in  Fig.~\ref{fig:framework}. The Soft Actor-Critic (SAC) algorithm serves as the baseline DRL method. To improve data efficiency, we employ a two-stage training strategy. Initially, a standard PID controller is initialized to generate imperfect trajectories for the DRL agent. Subsequently, the agent interacts directly with the simulation environment to update its policy. All transition data collected from the environment undergoes a state-based augmentation strategy before being stored into replay buffer. To mitigate the sim-to-real gap, we introduce a randomization method on the artificial muscle dynamics. In this paper, we denote the bootstrap method by letter `B', the augmentation method by letter `A', and the randomization of muscle dynamics by letter `R'. Thus, our proposed SAC algorithm with all three improvements is denoted by SAC-BAR.
	
	\subsection{Soft Actor-Critic}
	SAC \cite{SAC} is widely adopted as a strong baseline for continuous control tasks in robotics due to its robustness and sample efficiency. SAC incorporates an entropy-maximizing approach, with its learning objective defined as:
	\begin{equation}
		J(\pi)=\sum_{t=0}^T\mathbb{E}_{(s_t,a_t)\sim\rho_\pi}\left[\mathcal{R}(s_t,a_t)+\alpha\mathcal{H}(\pi(\cdot|s_t))\right],
	\end{equation}
	where $\mathbb{E}_{(s_t, a_t)\sim\rho_{\pi}}$ represents the expectation over the state-action pairs $(s_t,a_t)$ sampled from the policy distribution $\rho_{\pi}$, $\mathcal{H}(\pi(\cdot|s_t))$ represents the entropy of the policy $\pi$, and $\alpha$ serves as a temperature parameter. This entropy-regularized objective improves the robustness of DRL and accelerates its training \cite{DRL_survey}. We integrate a Gated Recurrent Unit (GRU) \cite{GRU} as the hidden layer in the actor-critic architecture to process sequential information.
	
	\subsection{Randomization of Muscle Dynamics}
	\begin{table}[t]
		\centering
		\caption{Dynamics parameters of randomization and their respective ranges in the robotic eye}
		\label{dr_param1}
		\begin{tabular}{l l l l}
			\toprule
			\textbf{Parameter }& \textbf{Value} & \textbf{Scailing factor} & \textbf{Additive term} \\
			\midrule
			$k$ $(\text{N}/\text{cm})$ & $0.25$ & $\mathrm{uniform}(0.8, 1.2)$ & -  \\
			$b$ $(\text{Ns}/\text{cm})$& $0.01$ & $\mathrm{uniform}(0.9, 1.1)$ & - \\
			$c$ $(\text{N}/^\circ\text{C})$& $0.0055$ & $\mathrm{uniform}(0.85, 1.15)$ & - \\
			$C_\text{th}$ $(\text{Ws}/^\circ\text{C})$ & $0.28$ & $\mathrm{uniform}(0.8, 1.2)$ & - \\
			$\lambda$ $(\text{W}/^\circ\text{C})$ & $0.094$ & $\mathrm{uniform}(0.85, 1.15)$ & - \\
			$R$ $(\Omega)$& $20.0$ & $\mathrm{uniform}(0.9, 1.1)$ & - \\
			$\theta, \psi$ noise $(^\circ)$ & - & - & $\mathrm{normal}(0, 0.1)$ \\
			$\dot{\theta}, \dot{\psi}$ noise $(^\circ/\text{s})$& - & - & $\mathrm{normal}(0, 0.05)$ \\
			\bottomrule
		\end{tabular}
	\end{table}
	\par
	The dynamic model of the artificial muscles considered in this paper is based on the literature\cite{SCP}, where the muscle's elasticity behavior is linearized to establish a dynamical system comprising a mechanical model characterized by stiffness and damping, given by
	\begin{equation}
		\label{eq:muscle-model1}
		F = k(x - x_0) + b\dot{x} + c(T(t)-T_\text{amb}),
	\end{equation}
	where $x$ and $x_0$ denote the current length and resting length of the actuator, respectively, while $k$, $b$, and $c$ denote the mean stiffness, damping, and temperature effect coefficients of the actuator, respectively. $T(t)$ denotes the current actuator temperature, while $T_\text{amb}$ stands for the ambient temperature of the environment. The thermoelectric model of the actuator is expressed as:
	\begin{equation}
		\label{eq:muscle-model2}
		C_\text{th}\frac{dT(t)}{dt} = P(t) - \lambda(T(t) - T_\text{amb}),
	\end{equation}
	where $C_\text{th}$ represents the thermal mass of the actuator (measured in $\text{Ws}/^\circ\text{C}$), $P(t)$ indicates the heat applied to the actuator, $\lambda$ is the absolute thermal conductivity of the actuator with respect to the ambient environment (in $\text{W}/^\circ\text{C}$). The applied power, $P(t)$, is controlled through the voltage applied to the muscle, expressed as $P(t)=V(t)^2/R$, where $R$ is the resistance of the artificial muscle. Equations (\ref{eq:muscle-model1}) and (\ref{eq:muscle-model2}) describe the dynamics of the artificial muscles. In the simulation, when the RL agent generates an action $a_t$, representing the voltages applied to the muscles, these equations are numerically solved to update the muscle states and compute their output forces. The computed forces are then incorporated into the dynamic model of the robotic system to update the overall system state ($s_t$).
	
	\begin{table}[t]
		\centering
		\caption{Dynamics parameters of randomization and their respective ranges in the robotic wrist}
		\label{dr_param2}
		\begin{tabular}{l l l l}
			\toprule
			\textbf{Parameter }& \textbf{Value} & \textbf{Scailing factor} & \textbf{Additive term} \\
			\midrule
			$k$ $(\text{N}/\text{cm})$ & $2.1$ & $\mathrm{uniform}(0.8, 1.2)$ & -  \\
			$b$ $(\text{Ns}/\text{cm})$& $0.63$ & $\mathrm{uniform}(0.9, 1.1)$ & - \\
			$c$ $(\text{N}/^\circ\text{C})$& $0.0707$ & $\mathrm{uniform}(0.85, 1.15)$ & - \\
			$C_\text{th}$ $(\text{Ws}/^\circ\text{C})$ & $3.06$ & $\mathrm{uniform}(0.8, 1.2)$ & - \\
			$\lambda$ $(\text{W}/^\circ\text{C})$ & $0.1189$ & $\mathrm{uniform}(0.85, 1.15)$ & - \\
			$R$ $(\Omega)$& $10.0$ & $\mathrm{uniform}(0.9, 1.1)$ & - \\
			$\theta, \phi$ noise $(^\circ)$ & - & - & $\mathrm{normal}(0, 0.1)$ \\
			$\dot{\theta}, \dot{\phi}$ noise $(^\circ/\text{s})$& - & - & $\mathrm{normal}(0, 0.05)$ \\
			\bottomrule
		\end{tabular}
	\end{table}
	
	The core idea behind dynamics randomization is to introduce randomness into specific dynamic parameters within the simulator during training. While this method is typically used in the field of manipulation and locomotion\cite{peng2018sim, xie2021dynamics}, its integration into the control of artificial muscle-driven robots is largely underexplored. We hypothesize that this method is particularly effective due to the unique uncertainties inherent in these systems. Artificial muscle actuators are subject to significant parameter variability caused by factors such as manufacturing tolerances, wear, and environmental conditions. By exposing the RL agent to a diverse range of dynamics during training, this approach facilitates the development of robust control policies that generalize effectively to real-world conditions with unmodeled dynamics.
	
	Specifically, a distribution $\rho_\mu$ of the muscle parameter set $\mu$ is established to encompass the variability of muscle dynamic parameters for dynamics randomization purposes. According to (\ref{eq:muscle-model1}) and (\ref{eq:muscle-model2}), the dynamic parameter set $\mu$ for randomization is given by $\mu:=\left\{k, b, c, C_\text{th}, \lambda, R\right\}$, where $:=$ is used to denote an assignment operation in this paper. Table~\ref{dr_param1} and \ref{dr_param2} outline the distribution $\rho_\mu$ of SCP and TCA muscles, respectively, along with the observation noise added for the robotic eye and wrist systems, where $\mathrm{uniform}(a, b)$ denotes a uniform distribution over interval $[a, b]$ and $\mathrm{normal}(a, b)$ denotes a normal distribution with mean $a$ and standard deviation $b$. The distribution $\rho_\mu$ is centered on the mean values derived from system identification to ensure alignment with the physical systems. The randomization ranges of these parameters are determined empirically to capture realistic variability while ensuring physical feasibility. These ranges are derived from standard deviations reported in prior studies on artificial muscle properties, including influences from material properties, environmental factors, and manufacturing differences\cite{SCP, TCA, zhang2017modeling}. Details on the impact of randomization variance on control performance are discussed in Section~\ref{subsec:range}. During the training process, the physical parameters of the actuators are randomized within predefined scaling ranges (i.e., $\mu\sim\rho_\mu$) at the start of each episode and remain constant throughout. Observation noise is sampled independently at each time step from a Gaussian distribution with a constant mean and variance, as specified in Table~\ref{dr_param1} and \ref{dr_param2}.
	
	\subsection{Bootstrap via PID}
	On-line DRL agents primarily rely on random exploration \cite{DRL_survey}. However, due to the slow response characteristics of artificial muscles, random exploration policies often fail to explore the state space comprehensively and efficiently, leading to low sample efficiency. To address this limitation, we accelerate the early-stage training by incorporating demonstration data into the initial replay buffer. For muscle-driven robotic systems, PID controllers are typically sufficient to approach target states and deliver moderate control performance. As such, they can generate trajectories that explore the defined state space more comprehensively, providing a beneficial starting point for the DRL learning process. Specifically, we initialize a PID controller that offers a barely acceptable (and typically not excellent) control performance before DRL training begins. Within the first $M$ episodes of training, the PID controller interacts with the environment and generates trajectories, which are then stored in the replay buffer along with the rewards. Compared to using explicit pre-training with PID data or a supervised loss for demonstration learning, our design mitigates the potential risk of the policy overfitting to suboptimal PID behavior before transitioning to RL.
	
	\subsection{Augmentation of State Vector}
	Data augmentation refers to a class of techniques aimed at increasing the size and improving the quality of training datasets\cite{Aug_survey}. The concept involves transforming limited data into diverse and functionally equivalent datasets. Given the labor-intensive and computationally expensive nature of collecting training data in robotic systems, we integrate data augmentation into the agent's learning process. This paper proposes a novel and straightforward state-based data augmentation technique designed for our targeted robotic systems.
	
	\begin{algorithm}[t]
		\caption{SAC-BAR Learning Control}
		\label{algorithm2}
		\begin{algorithmic}	[1]
			\STATE	Initialize critic networks $Q_{\theta_1},Q_{\theta_2}$, and actor network $\pi_{\phi}$ with random parameters $\theta_1,\theta_2,\phi$ 
			\STATE	Initialize target network parameters $\bar{\theta}_1 \leftarrow \theta_1,\bar{\theta}_2 \leftarrow \theta_2$ 
			\STATE	Initialize an empty replay buffer $\mathcal{B}$
			\STATE	Initialize a PID controller $\pi_{\text{PID}}$ 
			\STATE	Initialize dynamics parameters distribution $\rho_\mu$
			\FOR{$\text{episode}=1$ \TO $M$}
			\STATE Sample dynamics $\mu\sim\rho_\mu$
			\STATE Generate rollout $\tau_0=(s_0, a_0, r_0, ..., s_T)$ with dynamics $\mu$ and $a_t\sim\pi_{\text{PID}}(s_t)$
			\FOR{$i=1$ \TO $n$}
			\STATE	Create new rollouts $\tau_0\xrightarrow{AUG}\tau_i$
			\ENDFOR
			\STATE Store $\tau_0,\tau_1,...,\tau_n$ into $\mathcal{B}$
			\ENDFOR
			\STATE
			\FOR{$\text{episode}=M+1$ \TO $N$}
			\STATE Sample dynamics $\mu\sim\rho_\mu$
			\STATE Generate rollout $\tau_0=(s_0, a_0, r_0, ..., s_T)$ with dynamics $\mu$ and $a_t\sim\pi_\phi(s_t)$
			\FOR{$i=1$ \TO $n$}
			\STATE	Create new rollouts $\tau_0\xrightarrow{AUG}\tau_i$
			\ENDFOR
			\STATE Store $\tau_0,\tau_1,...,\tau_n$ into $\mathcal{B}$
			\FOR{$i=1$ \TO $k$}
			\STATE Sample a mini-batch of rollouts from $\mathcal{B}$
			\STATE Calculate the gradients $\nabla J_Q(\theta_i),\nabla J_\pi(\phi),\nabla J(\alpha)$
			\STATE Update network weights:
			\STATE $\theta_i \leftarrow \theta_i - \lambda_Q \nabla J_Q(\theta_i)$
			\STATE $\phi \leftarrow \phi - \lambda_\pi \nabla J_\pi(\phi)$
			\STATE $\bar{\theta}_i \leftarrow \tau \theta_i + (1-\tau) \theta_i$
			\STATE Adjust temperature parameter:
			\STATE $\alpha \leftarrow \lambda_\alpha \nabla J(\alpha)$
			\ENDFOR
			\ENDFOR
		\end{algorithmic}
	\end{algorithm}
	
	\par
	Our objective is to generate additional data transitions with different rewards. As described in Section~\ref{preliminaries}, the reward $r_t$ is calculated as:
	\begin{equation}
		\label{r_t_new}
		r_t=\mathcal{R}(\boldsymbol{x}_t, \boldsymbol{y}^*,a_t).
	\end{equation}
	Here, $\boldsymbol{x}_t$ and $a_t$ are directly related to system dynamics, whereas the target output vector $\boldsymbol{y}^*$ is manually set and changeable. Consequently, the proposed augmentation of state vectors proceeds as follows. At each time step $t$, we fix the motion state vectors $\boldsymbol{x}_t,\boldsymbol{x}_{t+1}$ and the action $a_t$, and vary the target vector $\boldsymbol{y}^*$. In this paper, we introduce a random additive term into $\boldsymbol{y}^*$ as follows
	\begin{equation}
		\boldsymbol{y}^*_{\text{new}}=\boldsymbol{y}^{*}\pm \delta\boldsymbol{Z}.
	\end{equation}
	Here, $\boldsymbol{y}^*_{\text{new}}$ denotes the new target vector. $\delta$ acts as a scalar factor to control the scope of augmentation, which largely depends on the defined range. $\boldsymbol{Z}$ is a random vector sharing identical dimensions with $\boldsymbol{y}^*$, wherein the elements of $\boldsymbol{Z}$ range from $0$ to $1$. Thus, we derive a pair of new states $s'_t$ and $s'_{t+1}$. After that, a new corresponding reward $r'_t$ is calculated using \eqref{r_t_new}. Following the above steps, the original transition tuple is transformed into a new one, i.e.,
	\begin{equation}
		(s_t,a_t,r_t,s_{t+1})\xrightarrow{AUG}(s'_t,a_t,r'_t,s'_{t+1}),
	\end{equation}
	which creates different transition tuples at a negligible computational cost. We choose to perturb the target vectors instead of the state vectors because perturbing state vectors may introduce physical inconsistencies, which can prevent the agent from learning accurate dynamics. The pseudo code of SAC-BAR learning control framework is presented in Algorithm~\ref{algorithm2}.
	
	\begin{figure*}[t]
		\centering
		\subfloat{
			\includegraphics[width=\linewidth]{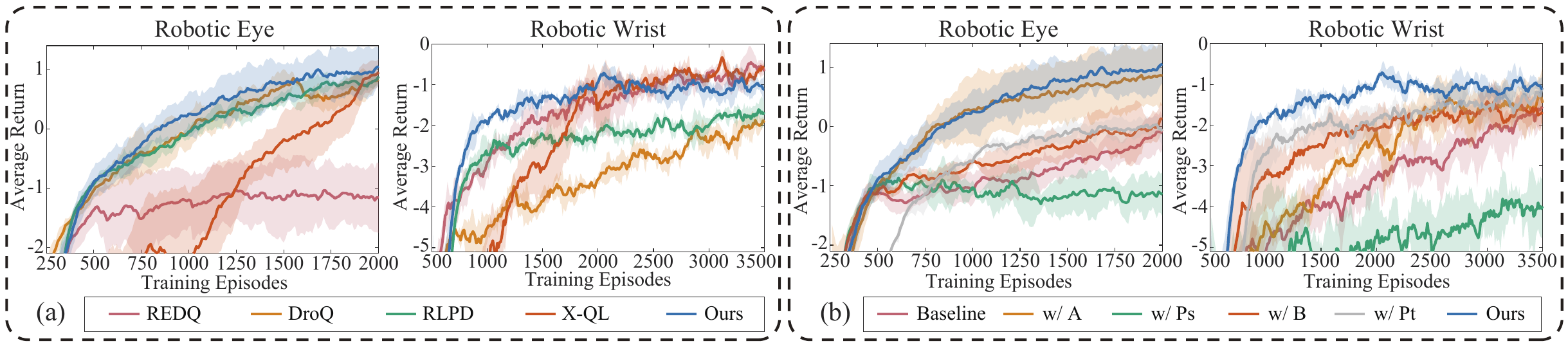}
		}		
		\vspace*{-0.1in}
		\caption{Reward trajectories with respect to the training episode number. The shaded regions indicate the standard deviation of the rewards across multiple runs. (a) Comparative results between our proposed SAC-BAR method and SOTA approaches, including REDQ, DroQ, RLPD, and X-QL. (b) Comparative results from the ablation study, where `A', `Ps', `B', and `Pt' represent the augmentation method, state perturbation variation, bootstrap mechanism, and pre-training variation, respectively.}
		\label{fig:reward-res}
	\end{figure*}
	
	\section{Simulation Experiments}
	\label{experiment in simulation}
	This section presents the training results from simulation environments, comparing our proposed SAC-BAR against several SOTA reinforcement learning algorithms in terms of sample efficiency on two robotic systems. Additionally, we perform ablation studies to evaluate the effectiveness of the proposed methods and discuss the choice of hyperparameters.
	
	\subsection{Training Setup}
	In each training episode, the robotic agent aims to reach a target state randomly selected from a uniformly distributed range. The initial PID controller parameters are set as $K_p = 2.1$, $K_i = 0.2$, and $K_d = 0.5$ for the robotic eye, and $K_p = 3.3$, $K_i = 0.5$, and $K_d = 0.3$ for the robotic wrist. The number of data augmentation is set as $n=10$. The neural network architecture for both the actor and critic consists of a fully connected input layer, a GRU layer with 256 hidden units, and a fully connected output layer, with ReLU as the activation function. The model is trained using a batch size of 20 trajectories and an experience replay buffer with a capacity of $1 \times 10^5$ trajectories. The learning rate is set to $3 \times 10^{-4}$. Additionally, the SAC temperature parameter $\alpha$ is configured to be auto-tuned, as described in\cite{SAC}. The hyperparameters $N$ and $M$ are empirically determined based on the system's complexity. Specifically, $N = 2000$ and $M = 250$ are used for the robotic eye, while $N = 3500$ and $M = 500$ are used for the robotic wrist. When the bootstrap method is not applied, random exploration is employed during the initial $N$ warm-up episodes. The algorithms are implemented in PyTorch and trained on a computer with an Intel Xeon Platinum 8383C CPU and an NVIDIA GeForce RTX 4090 GPU.
	
	\subsection{Comparison Results Against SOTA Methods}
	\par
	Several SOTA sample-efficient RL algorithms, tailored for state-based observations and continuous action spaces, are selected for comparison. These include REDQ\cite{REDQ}, DroQ\cite{DroQ}, RLPD\cite{RLPD}, and X-QL\cite{XQL}, which represent diverse approaches to improving sample efficiency. We train SAC-BAR on two robotic systems and compare its performance with these SOTA approaches. Performance is measured by the online per-step average reward achieved by the agents, with each algorithm trained across four independent runs. Fig.~\hyperref[fig:reward-res]{\ref*{fig:reward-res}(a)} presents the comparative results between SAC-BAR and the selected SOTA RL methods. In the robotic eye system, SAC-BAR achieves performance comparable to (and slightly better than) DroQ and RLPD, while significantly outperforming REDQ and X-QL in terms of sample efficiency. In the robotic wrist system, SAC-BAR not only significantly outperforms RLPD and DroQ but also demonstrates moderate improvements in convergence speed over REDQ and X-QL. Notably, REDQ's performance relies on a high UTD ratio and an ensemble of Q-functions, resulting in high computational cost; for instance, REDQ requires 18.2 hours for 3,500 training episodes, whereas SAC-BAR completes the same training in only 4.1 hours under our experimental setup. Moreover, X-QL exhibits significant instability during the early stages of training in both robotic systems. These results demonstrate that SAC-BAR is better suited than the existing SOTA methods for the string-type artificial muscle robots targeted in this study.
	
	\subsection{Ablation Study in Simulation}
	\par
	To evaluate the effectiveness of the proposed bootstrap and augmentation techniques, an ablation study is conducted on two robotic systems. Five variants of our proposed framework are tested as follows: (1)~\textbf{Baseline:} This variant employs only the SAC algorithm without any additional components. (2)~\textbf{w/~A:} This variant incorporates the augmentation method. (3)~\textbf{w/~Ps:} This variant perturbs the state vectors instead of target vectors. (4)~\textbf{w/ B:} This variant integrates the bootstrap mechanism. (5)~\textbf{w/~Pt:} This variant pre-trains the networks on data generated by the PID controller before exploration.
	\par
	Training results of these variants and our proposed SAC-BAR are presented in Fig.~\hyperref[fig:reward-res]{\ref*{fig:reward-res}(b)}. As illustrated in the figures, our proposed algorithm exhibits the highest level of data efficiency in both robotic systems. Compared to the baseline SAC algorithm, the integration of bootstrap and augmentation enhances data efficiency and expedites the training process. In the robotic eye system, the baseline algorithm achieves an average reward of $-0.10$ after $2,000$ training episodes, whereas our approach achieves comparable performance in only approximately $750$ episodes, resulting in a $2.67$-fold increase in data efficiency compared to the original level. Similarly, in the robotic wrist system, the baseline algorithm achieves an average reward of $-1.61$ after $3,500$ episodes, while ours achieves comparable performance in only about $1,200$ episodes, leading to a $2.92$-fold increase in data efficiency. The performance of the variant perturbing state vectors is lower compared to the baseline. We attribute this to the perturbation of state variables, which likely hinders the agent's ability to accurately learn the system dynamics. While pre-training on PID-generated data enhances training efficiency, we observe that overfitting may occur during the pre-training phase, leading to a significant drop in the average return.
	
	\subsection{Influence of Hyperparameters}
	\begin{figure}[t]
		\centering
		\includegraphics[width=\linewidth]{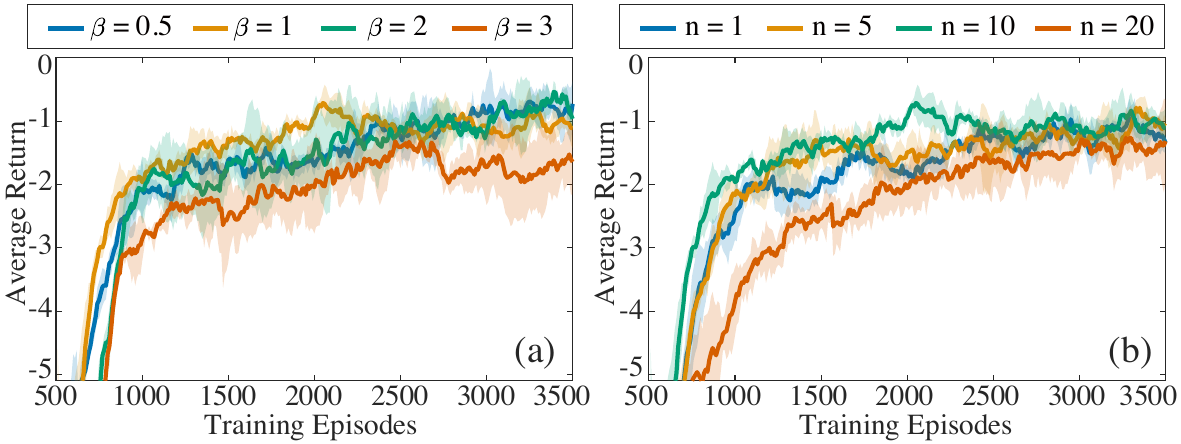}
		\vspace{-0.2in}
		\caption{(a) Comparative results for different scaling factors of the initial PID gains ($\beta = 0.5, 1.0, 2.0, 3.0$), where $K'_p = \beta K_p$, $K'_i = \beta K_i$, and $K'_d = \beta K_d$. (b) Comparative results for different numbers of augmentation samples ($n = 1, 5, 10, 20$).}
		\label{fig:exp-pid-n}
	\end{figure}
	The hyperparameters that significantly affect the performance of SAC-BAR include the initial PID gains and the number of augmentation samples. To evaluate their impact, we train the algorithm under different hyperparameter settings on the robotic wrist. Specifically, we compare the training results using different scaling factors for the initial PID gains ($\beta = 0.5, 1.0, 2.0, 3.0$) and varying numbers of augmentation samples ($n = 1, 5, 10, 20$), as shown in Fig.~\ref{fig:exp-pid-n}. The results indicate that our method is generally robust to these hyperparameters. When the PID gains are within a reasonable range (not necessarily optimal), the differences in training results are minor. However, when the PID gains are extreme (e.g., $\beta = 3$), the demonstrations it provides are of poor quality, thus affecting the algorithm's training performance. Similarly, the performance of the model trained with $n = 20$ drops significantly, as excessive augmentation introduces redundancy, causing the agent to overfit to the augmented data rather than focusing on the true dynamics of the environment. On the other hand, an extremely low $n$ reduces the diversity of the training data, slightly degrading performance. A moderate value of $n$ (e.g., $n=5$ or $n=10$) strikes a balance between data diversity and training stability.

	\section{Real-World Experiments}
	\label{real world experiments}
	This section presents the real-world experimental results of the proposed SAC-BAR. We first demonstrate the results of an ablation study conducted on the robotic wrist platform, which validate the effectiveness of the proposed dynamics randomization in enhancing control performance in real-world scenarios. Next, experimental results on the robotic eye are provided to show the generalizability of SAC-BAR across similar robotic systems. Additionally, we investigate the impact of the randomization variance on control performance.
	
	\subsection{Comparison Results on Robotic Wrist}	
	\begin{figure}[t]
		\centering
		\includegraphics[width=\linewidth]{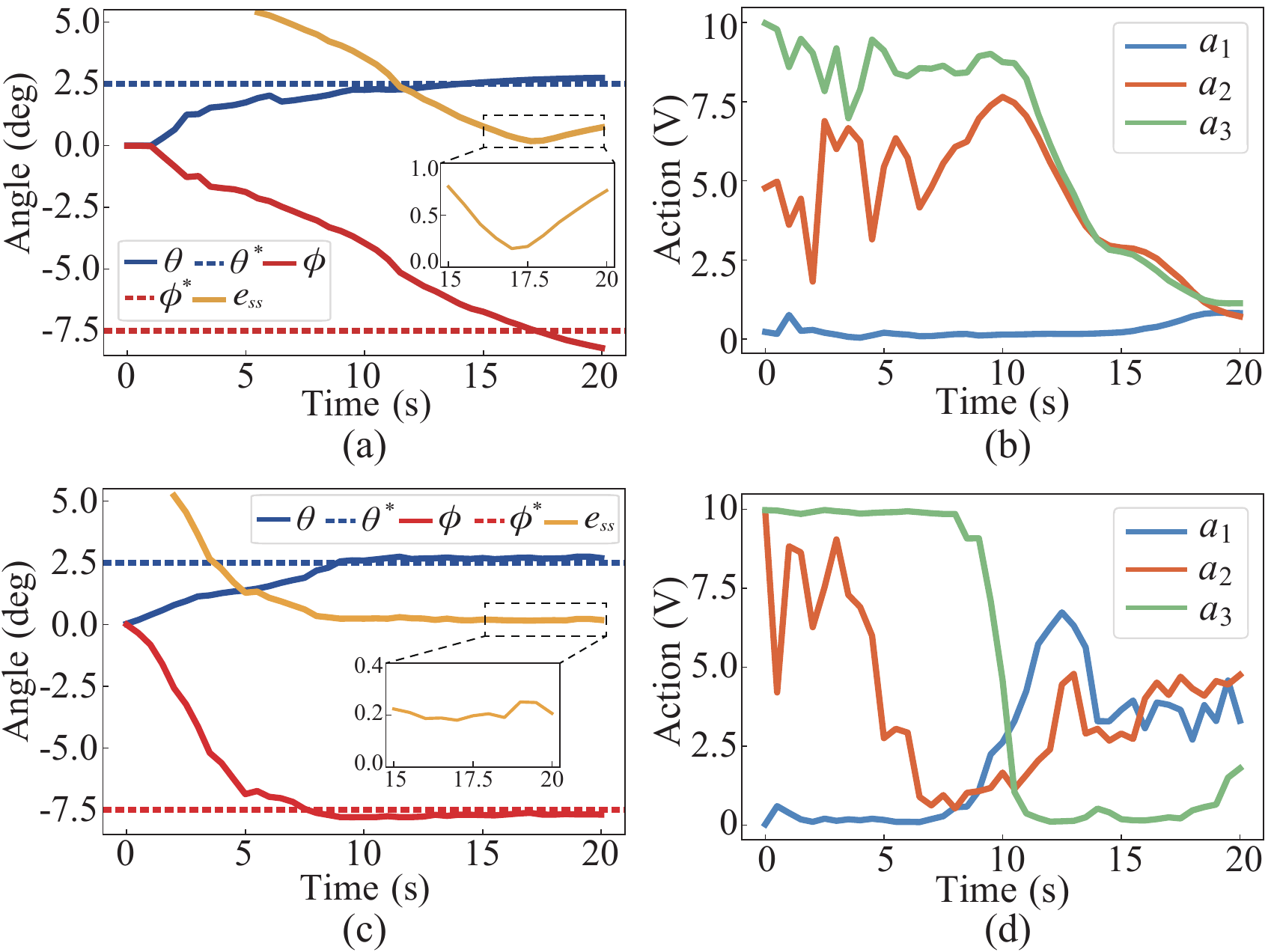}
		\vspace{-0.2in}
		\caption{Comparison of experimental trajectories between the baseline and SAC-BAR. Target angles are set at ($\theta^*=2.5^\circ,\phi^*=-7.5^\circ$). (a), (b) The system output angles ($\theta, \phi, e_\text{ss}$) and corresponding actions $a_t$ of the baseline SAC algorithm. (c), (d) Those of our proposed SAC-BAR algorithm.}
		\label{fig:wrist-com2}
	\end{figure}
	
	To evaluate the impact of each component on real-world control performances, three different control policies are tested on the robotic wrist platform, including the vanilla SAC algorithm, SAC with only randomization of muscles (SAC-R), and our proposed SAC-BAR. The parallel robotic wrist starts from the initial state with $\theta=\phi=0^\circ$ and attempts to reach and maintain the target angle set $(\theta^*,\phi^*)$. The action vector $a$ is computed by the trained actors at a time step of $t_a=0.5$\,s, and each episode spans a duration of $t_\text{total}=20$\,s. In Fig.~\hyperref[fig:wrist-com2]{\ref*{fig:wrist-com2}(a)}, it is evident that without randomization of muscle dynamics, the SAC agent struggles to achieve the desired orientation. While the agent may eventually reach the desired orientation at certain target angles, maintaining that target angle becomes a challenge, resulting in a considerable steady-state error. Conversely, as observed in Fig.~\hyperref[fig:wrist-com2]{\ref*{fig:wrist-com2}(c)}, the proposed SAC-BAR agent achieves the target state in less than $10$\,s and consistently remains within its neighborhood thereafter. As illustrated by the input action trajectories in Fig.~\hyperref[fig:wrist-com2]{\ref*{fig:wrist-com2}(d)}, this success is attributed to the agent learning to maximize the action value during the rising phase and promptly reduce it upon approaching the target angles. Conversely, in Fig.~\hyperref[fig:wrist-com2]{\ref*{fig:wrist-com2}(b)}, there exist serious oscillations in the action trajectories of SAC during the initial $10$\,s, leading to a prolonged rise time and sometimes even failure to reach the target.
	
	\begin{figure*}[t]
		\centering
		\includegraphics[width=\linewidth]{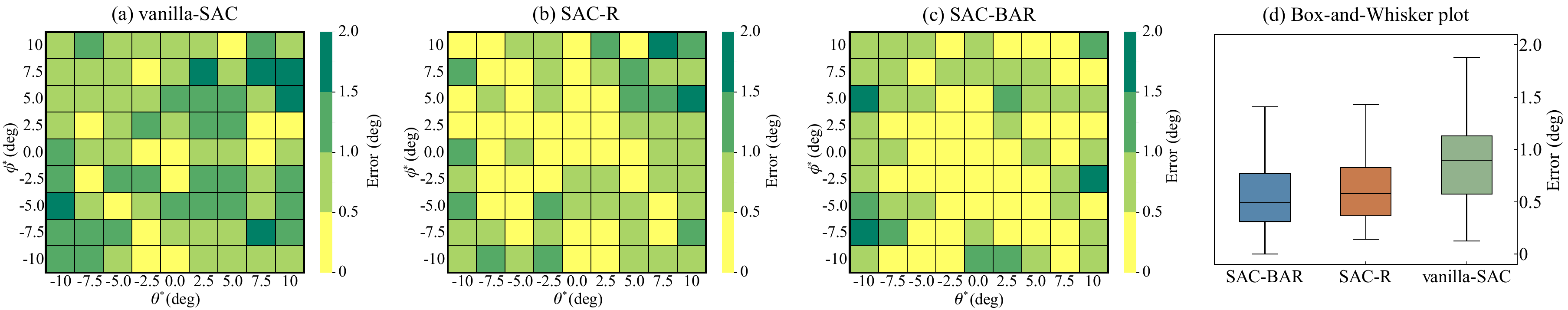}
		\caption{Experimental results of the steady-state error tests in the parallel robotic wrist under selected control algorithms. (a), (b), (c) Heat maps illustrating the experimental results for the SAC, SAC-R, and SAC-BAR, respectively. The combination of target angles $(\theta^*, \phi^*)$ is systematically spaced at $2.5^\circ$ intervals within the range $(|\theta^*|,|\phi^*|)\leq 10^\circ$. The color bars in the visualization indicate the Euclidean distance norm of steady-state errors. (d) The box charts illustrating the medians, maxima, minima, and quartiles of the steady-state errors for the selected comparative control algorithms.}
		\label{fig:wrist_sse}
	\end{figure*}
	
	\par
	To comprehensively evaluate the control performance of the algorithms across different target angles, we design a steady-state error field test for the robotic wrist. We select a series of target angles $(\theta^*,\phi^*)$ spaced at $2.5^\circ$ intervals within the defined range $\theta^*,\phi^*\in[-10^\circ,10^\circ]$. Each agent is set to run for $25$\,s, sequentially targeting each angle pair. The steady-state error is measured and averaged over the last $5$s window, determined by computing the Euclidean distance norm between the current and target angles, defined as $e_\text{ss}=\sqrt{(\theta-\theta^*)^2+(\phi-\phi^*)^2}$. The steady-state errors for different targets of SAC, SAC-R, and SAC-BAR are presented in Fig.~\hyperref[fig:wrist_sse]{\ref*{fig:wrist_sse}(a)}, \hyperref[fig:wrist_sse]{(b)}, and \hyperref[fig:wrist_sse]{(c)}, respectively. The $\mathrm{x}$ and $\mathrm{y}$ axes therein represent $\theta^*$ and $\phi^*$, while the color bars represent the averaged steady-state errors. Following that, box charts are illustrated in Fig.~\hyperref[fig:wrist_sse]{\ref*{fig:wrist_sse}(d)}. These figures demonstrate that SAC agent underperforms across nearly the entire defined range, whereas both SAC-R and SAC-BAR show significantly improved control performance. These results indicate that the proposed randomization of muscle dynamics significantly enhances the real-world control performance. 
	\begin{table}[t]
		\centering
		\caption{The simulation (Sim) and experimental (Real) results of steady state error using different control methods}
		\label{table_sse}
		\renewcommand{\arraystretch}{1.15}
		\setlength{\tabcolsep}{2.9pt}
		\begin{tabular}{l c c c c c c}
			\toprule
			\multirow{2}{0.2in}{$\bar{e}_\text{ss}$ (deg)} & \multicolumn{6}{c}{Control Method} \\
			\cmidrule(l{1.7em}r{1.2em}){2-7}
			& \textbf{PID} & \textbf{AC} & \textbf{PPO} & \textbf{SAC} & \textbf{SAC-R} & \textbf{SAC-BAR} \\
			\midrule
			\textbf{Sim} & 1.12{\tiny $\pm0.47$} & 0.80{\tiny $\pm0.41$} & 0.79{\tiny $\pm0.32$} & 0.46{\tiny $\pm0.27$} & 0.47{\tiny $\pm0.33$} & 0.40{\tiny $\pm0.17$} \\
			\textbf{Real} & 1.03{\tiny $\pm0.49$} & 0.86{\tiny $\pm0.67$} & 1.24{\tiny $\pm0.48$} & 0.91{\tiny $\pm0.46$} & 0.66{\tiny $\pm0.39$} & 0.58{\tiny $\pm0.37$} \\
			\bottomrule
		\end{tabular}
	\end{table}
	\par
	To evaluate the real-world performance of our method against traditional control approaches, we implement the initial PID controller, an adaptive controller (AC), and the Proximal Policy Optimization (PPO) algorithm\cite{schulman2017proximal} for comparison. The steady-state errors averaged over the defined range, denoted as $\bar{e}_\text{ss}$, for these methods are tested in both simulation and real-world scenarios. The experimental results are summarized in Table~\ref{table_sse}. In real-world experiments, our method achieves superior performance with a mean error of $0.58^\circ$, significantly lower than those of the other methods. Specifically, compared to PID, AC, PPO, and SAC, our method reduces the mean error by $43.7\%$, $32.6\%$, $53.2\%$, and $36.3\%$, respectively. While PPO and SAC are strong DRL baselines, they struggle with the sim-to-real gap, resulting in larger errors and less stable behaviors. The adaptive controller performs better due to its robustness to parameter uncertainties, but it requires extensive parameter tuning and still exhibits suboptimal performance at certain target angles. A comparison of SAC and SAC-R results demonstrates that the randomization of muscle dynamics effectively bridges the sim-to-real gap, leading to reduced disparity between simulated and actual results. Furthermore, SAC-BAR achieves more comprehensive training in simulation than SAC-R, resulting in a slight improvement in control performance.
	
	\subsection{Learning Control of Robotic Eye}
	
	\begin{figure}[t]
		\centering
		\includegraphics[width=\linewidth]{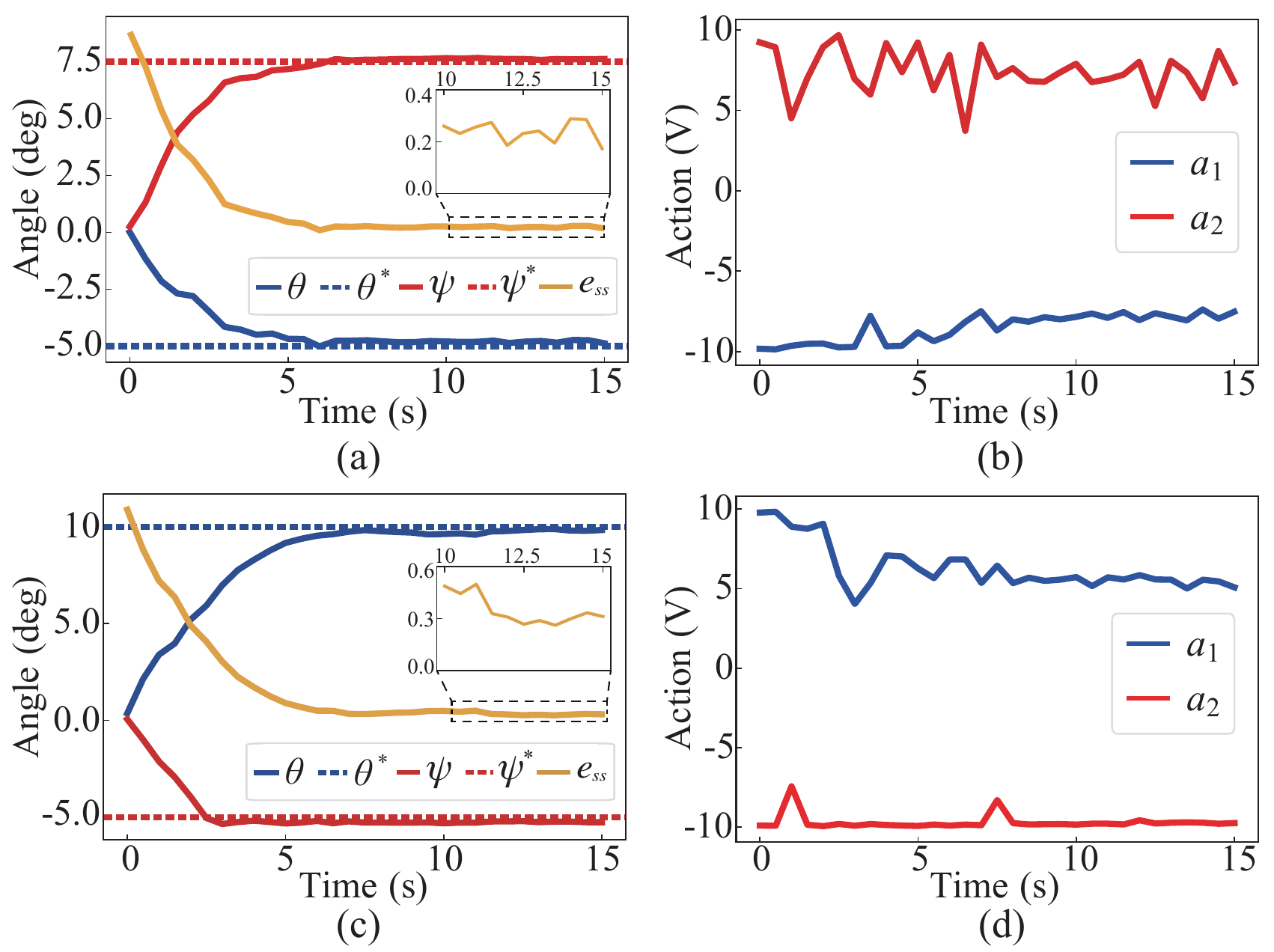}
		\vspace{-0.20in}
		\caption{Experimental trajectories of the foveation test of the robotic eye. Target angles are set at $(\theta^*=-5^\circ,\psi^*=7.5^\circ)$ and $(\theta^*=10^\circ,\psi^*=-5^\circ)$, respectively. (a), (c) The system output angles $(\theta,\psi,e_\text{ss})$. (b), (d) The corresponding actions $a_t$ generated by the trained actor.}
		\label{fig:eye_traj}
	\end{figure}
	
\begin{figure}[t]
		\centering
		\includegraphics[width=0.50\linewidth]{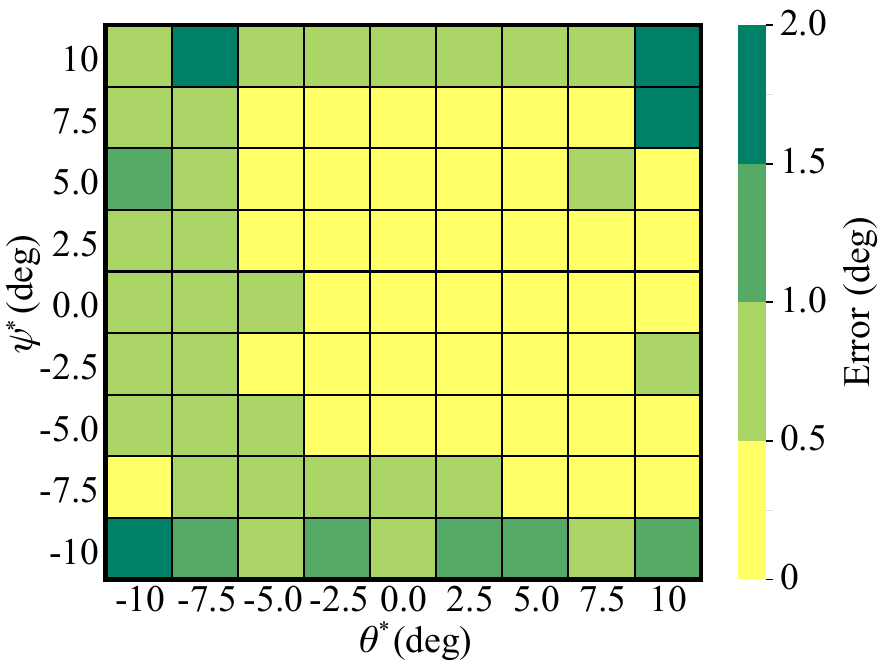}
		\caption{Experimental results on the steady-state errors in robotic eye at eyery combination of $(\theta^*,\psi^*)$ target angles spaced at $2.5^\circ$ intervals for $(|\theta^*|,|\psi^*|)\leq 10^\circ$. }
		\label{fig:eye_sse}
	\end{figure}

	To demonstrate the applicability of the proposed learning control framework across diverse robotic systems driven by artificial muscle strings, this subsection presents experimental results on an additional platform. The robotic eye, initialized at $\theta=\psi=0^\circ$ at ambient room temperature, is configured to foveate at various target angles $(\theta^*,\psi^*)$. Considering the system's rise time, each testing episode lasts for $t_\text{total}=15$\,s, with an action time step of $t_a=0.5$\,s.  The averaged Euclidean distance  between the current and target angles within the last $5s$ window serves as the steady-state error metric, i.e., $e_\text{ss}=\sqrt{(\theta-\theta^*)^2+(\psi-\psi^*)^2}$. Experimental trajectories with targets set at $(\theta^*,\psi^*)=(-5^\circ,7.5^\circ)$ and $(10^\circ,-5^\circ)$ are illustrated in Fig~\ref{fig:eye_traj}. These trajectories demonstrate that the measured angles rise to the target in approximately $5$\,s, with negligible steady-state error. A similar field test is conducted to evaluate the overall control performance of the algorithm. The results, illustrated in Fig.~\ref{fig:eye_sse}, reveal solid control performance, with steady-state errors consistently below $1^\circ$ for most target angles and an average of $0.56^\circ$.  Our prior research\cite{Eye_RAL} employed a Deep Deterministic Policy Gradient-based controller\cite{lillicrap2015continuous}, which was trained in simulation for $50,000$ episodes. In this paper, our proposed algorithm achieves comparable control performance with only $2,000$ training episodes, representing just $1/25$ of the original training time. 
	
	\subsection{Different Randomization Variances}
	\label{subsec:range}
	\begin{table}[t]
		\centering
		\caption{Comparison of control errors for models trained with different variances of the randomization distribution}
		\label{tab:diff-range}
		\renewcommand{\arraystretch}{1.15}
		\setlength{\tabcolsep}{8.5pt}
		\begin{tabular}{l c c c c}
			\toprule
			\multirow{2}{0.4in}{$\bar{e}_\text{ss}$ (deg)} & \multicolumn{4}{c}{Standard Deviation Values} \\
			\cmidrule(l{1.2em}r{0.6em}){2-5}
			& $0.5\sigma_0$ & $1.0\sigma_0$ & $1.5\sigma_0$ & $2.0\sigma_0$ \\
			\midrule
			\textbf{Sim} & 0.38{\tiny $\pm0.17$} & 0.40{\tiny $\pm0.17$} & 0.60{\tiny $\pm0.35$} & 0.74{\tiny $\pm0.41$} \\
			\textbf{Real} & 0.63{\tiny $\pm0.49$} & 0.58{\tiny $\pm0.37$} & 0.64{\tiny $\pm0.38$} & 0.76{\tiny $\pm0.43$} \\
			\bottomrule
		\end{tabular}
	\end{table}
	\par
	To evaluate the influence of randomization variance on control performance, we conduct experiments using four standard deviation (SD) settings: $0.5\sigma_0$, $1.0\sigma_0$, $1.5\sigma_0$, and $2.0\sigma_0$, where $\sigma_0$ represents the baseline SD. Controllers trained under each randomization setting are evaluated in both simulated and real-world environments using the robotic wrist platform. As summarized in Table~\ref{tab:diff-range}, the controller trained with a $2.0\sigma_0$ range exhibits a significant degradation in performance across both domains. In contrast, the controller trained with a $0.5\sigma_0$ range demonstrates limited generalization, failing to accommodate the variability inherent in real-world dynamics. While the selection of randomization variance significantly affects the resultant control performance, all  controllers successfully converge during training. This indicates that the proposed method ensures reliable convergence across a reasonably bounded range of randomization settings. These results underscore the necessity of carefully tuning the randomization variance to balance generalization and performance without compromising convergence stability.
	
	\section{Conclusion}
	\label{conclusions}
	This paper presents a learning control framework SAC-BAR designed for the string-type artificial muscle-driven robotic systems. To enhance training data efficiency, we proposed two methods, namely bootstrap via PID and state-based data augmentation. Additionally, we incorporated randomization of artificial muscle dynamic parameters during training to address the sim-to-real gap for enhancing real-world control performance. Extensive experiments and ablation studies on two robotic systems---a robotic eye and a parallel robotic wrist---validated the effectiveness and generalizability of the proposed control framework. In simulation, the designed bootstrap and augmentation methods reduced the required training episodes by approximately 60\% compared to the baseline algorithm. Real-world experiments demonstrated that randomization of muscle dynamics improved the control performance in targeted robotic systems, reducing the average error by 36.3\%. Furthermore, our framework demonstrated satisfactory performance on both robotic platforms, highlighting its potential applicability to other robotic systems with actuators characterized by high nonlinearity and parameter uncertainties.
	\par
	Despite the strong performance of the proposed SAC-BAR, several limitations exist. The configuration of the reward function relies on empirical tuning, and this paper does not provide a formal theoretical convergence analysis under muscle dynamic randomization. Additionally, the application scenarios explored in this work are relatively limited, and the robotic testing platforms utilized share similar characteristics. In future work, we aim to explore automated methods for reward function tuning and develop theoretical convergence analysis frameworks that consider bounded randomization ranges and non-stationary dynamics. To further evaluate the practicality of the proposed method, we plan to extend its application to more complex systems and diverse scenarios.
	
	\bibliographystyle{IEEEtran}
	%	\footnotesize
	\bibliography{ref}

\end{document}